\documentclass[twoside]{article}

\usepackage[accepted]{aistats2019} 

\usepackage{srcltx} 
\usepackage{amssymb,amsmath,bm,paralist,booktabs,color,amsthm,enumitem,url,soul}
\usepackage[mathscr]{eucal}

\newcommand{\tb}{\textbf}   
\renewcommand{\b}{\mathbf}  
\newcommand{\R}{\mathbb{R}}  
\newcommand{\N}{\mathbb{N}}  
\renewcommand{\d}{\mathrm{d}} 
\renewcommand{\H}{\mathscr{H}} 
\newcommand{\bo}{\bm{\omega}} 
\newcommand{\p}{\partial}
\newcommand{\K}{\mathscr{S}}
\renewcommand{\S}{\Lambda}
\renewcommand{\P}{\mathbb{P}} 
\newcommand{\Mp}[1]{\mathscr{M}^1_+\left(#1\right)} 
\newcommand{\F}{\mathcal{F}}
\newcommand{\G}{\mathcal{G}}
\renewcommand{\O}{O} 
\renewcommand{\o}{o} 
\newcommand{\vol}{\text{vol}} 
\newcommand{\X}{\mathscr{X}}

\usepackage{mathtools}
\DeclarePairedDelimiter{\ceil}{\lceil}{\rceil}

\newtheorem*{theorem}{Theorem}

\begin{document}


\twocolumn[
\aistatstitle{On Kernel Derivative Approximation with Random Fourier Features}
  
 \aistatsauthor{Zolt{\'a}n Szab{\'o} \And Bharath K. Sriperumbudur}
 \aistatsaddress{CMAP, {\'E}cole Polytechnique\\ 
 Route de Saclay, 91128 Palaiseau, France\\
 \texttt{zoltan.szabo@polytechnique.edu}
 \And  Department of Statistics\\
      Pennsylvania State University\\
      314 Thomas Building\\ 
      University Park, PA 16802\\
      \texttt{bks18@psu.edu}}
]

\begin{abstract}
Random Fourier features (RFF) represent one of the most popular and wide-spread techniques in machine learning to scale up kernel algorithms. 
Despite the numerous successful applications of RFFs, unfortunately, quite little is understood theoretically on their optimality and  limitations of their performance.
Only recently, precise statistical-computational trade-offs have been established for RFFs in the approximation of kernel values,  kernel ridge regression, kernel PCA and SVM classification.
Our goal is to spark the investigation of optimality of RFF-based approximations in tasks involving not only function values but \emph{derivatives}, which naturally lead to 
optimization problems with kernel derivatives.
Particularly, in this paper, we focus on the approximation quality of RFFs for kernel derivatives and prove that the existing finite-sample guarantees can be 
improved exponentially in terms of the domain where they hold, using recent tools from unbounded empirical process theory. Our result implies that 
the same approximation guarantee is attainable for kernel derivatives using RFF as achieved for kernel values.
\end{abstract} 

\section{INTRODUCTION}
Kernel techniques \cite{berlinet04reproducing,steinwart08support,paulsen16introduction} are among the most influential and widely-applied tools, with significant impact on virtually all areas of 
machine learning and statistics. Their versatility stems from the function class associated to a kernel called reproducing kernel Hilbert space (RKHS) \cite{aronszajn50theory} which shows tremendous success in modelling complex relations.

The key property that makes kernel methods computationally feasible and the optimization over RKHS tractable is the representer theorem \cite{kimeldorf71some,scholkopf01generalized,yu13characterizing}. 
Particularly, given samples $\{(x_i,y_i)\}_{i=1}^n \subset \X \times \R$, consider the regularized empirical risk minimization problem specified by a kernel $k: \X \times \X \rightarrow \R$, the 
associated RKHS $\H_k\subset \R^{\X}$, a loss function $V: \R \times \R \rightarrow \R^{\ge 0}$, and a penalty parameter $\lambda >0$:
\begin{align}
 \min_{f\in \H_k}J_0(f) &:= \frac{1}{n}\sum_{i=1}^n V(y_i,f(x_i)) + \lambda \left\|f\right\|_{\H_k}^2, \label{eq:J}
\end{align}
where 
$\H_k$ is the Hilbert space defined by the following two properties: 
\begin{enumerate}[labelindent=0cm,leftmargin=*,topsep=0cm,partopsep=0cm,parsep=0cm,itemsep=0.2cm]
  \item $k(\cdot,x) \in \H_k$ ($\forall x\in\X$),\footnote{$k(\cdot,x)$ denotes the function $y\in \X \mapsto k(y,x) \in \R$ while keeping $x\in \X$ fixed.} and 
  \item $f(x)=\left<f,k(\cdot,x)\right>_{\H_k} $ $(\forall x\in \X, \forall f\in \H_k)$, which is called the reproducing property.
\end{enumerate}

Examples falling under \eqref{eq:J} include e.g., kernel ridge regression with the  squared loss or soft-classification with the hinge loss:
\begin{align*}
 V(f(x_i),y_i) &= (f(x_i)-y_i)^2,\\
 V(f(x_i),y_i) &= \max(1-y_if(x_i),0).
\end{align*}

\eqref{eq:J} is an optimization problem over a function class ($\H_k$) which could generally be intractable. Thanks to the specific structure of RKHS, however, the representer theorem enables one to parameterize
the optimal solution of \eqref{eq:J} by finitely many coefficients:
\begin{align}
 f(\cdot) = \sum_{j=1}^n c_j k(\cdot,x_j), \quad c_j \in \R.
\end{align}
As a result, \eqref{eq:J} becomes a finite-dimensional optimization problem determined by the \emph{pairwise similarities}  of the samples [$k(x_i,x_j)$]:
\begin{align}
 \min_{\b c\in \R^n}\tilde{J}_0(\b c) &:= \frac{1}{n}\sum_{i=1}^n V\left(y_i,\sum_{j=1}^n c_j k(x_i,x_j)\right)\nonumber \\
      &  \qquad +\lambda \sum_{i=1}^n \sum_{j=1}^n c_i c_j k(x_i,x_j), \label{eq:k(x,y)}
\end{align}
where the second term follows from the reproducing property of kernels.

However, in many learning problems such as nonlinear variable selection \cite{rosasco10regularization,rosasco13nonparametric}, (multi-task) gradient learning \cite{ying12learning}, semi-supervised or Hermite learning with gradient information \cite{zhou08derivative, shi10hermite}, or density estimation with  infinite-dimensional exponential families \cite{sriperumbudur17density}, apparently considering the \emph{derivative information} ($\p^{\b p}f(\b x_i):= \frac{\p^{p_1+\ldots+p_d}f(\b x_i)}{\p_{x_1}^{p_1}\cdots \p_{x_d}^{p_d}}$, $\X := \R^d$) other than just the function values ($f(\b x_i)$) turns out to be \emph{beneficial.}
In these tasks containing derivatives, \eqref{eq:J} is generalized with loss functions $V_i:  \R^{|I_i|+1} \rightarrow \R^{\ge 0}$ ($i=1,\ldots,n$; $|I_i|$ denotes the cardinality of $I_i$) to the form
\begin{align}
 \min_{f\in \H_k} J(f)\hspace*{-0.04cm} &:= \hspace*{-0.05cm} \frac{1}{n}\sum_{i=1}^n V_i\left(y_i,\left\{\p^{\b p} f(\b x_i)\right\}_{\b p\in I_i}\right)\hspace*{-0.03cm} + \hspace*{-0.03cm}\lambda \left\|f\right\|_{\H_k}^2.  \label{eq:J2}
\end{align}
The solution of this minimization task ---similar to \eqref{eq:J}---enjoys a finite-dimensional parameterization \cite{zhou08derivative}:
\begin{align*}
    f(\cdot) & = \sum_{j=1}^n \sum_{\b p \in I_j} c_{j,\b p}\p^{\b p, \b 0}k(\cdot, \b x_j), \quad (c_{j,\b p} \in \R),
\end{align*}
where $\p^{\b{p},\b{q}}k(\b{x},\b{y}) := \frac{\partial^{\sum_{i=1}^d (p_i + q_i)} k(\b{x},\b{y})}{\partial^{p_1}_{x_1}\cdots\partial^{p_d}_{x_d} \partial^{q_1}_{y_1}\cdots\partial^{q_d}_{y_d}}$. Hence, the 
optimization in \eqref{eq:J2} can be reduced to 
\begin{eqnarray}
  \lefteqn{\min_{\b c} \tilde{J}\left( \b c\right)  =} \nonumber\\
     &&\hspace*{-.3cm} = \frac{1}{n}\sum_{i=1}^n V_i\left(y_i,\Bigg\{ \sum_{j=1}^n \sum_{\b p \in I_j} c_{j,\b p}\p^{\b p, \b 0}k(\b x_i, \b x_j) \Bigg\}_{\b p\in I_i}\right) \nonumber\\
     && \hspace*{-.3cm}\quad +\lambda \sum_{i=1}^n \sum_{\b p \in I_i} \sum_{j=1}^n \sum_{\b q \in I_j} c_{i,\b p} c_{j,\b q} \p^{\b p, \b q}k(\b x_i, \b x_j), \label{eq:kder-finiteD-opt}
\end{eqnarray}
where $\b c = \left( c_{i,\b p}\right)_{i\in \{1,\ldots,n\},\b p \in I_i} \in \R^{\sum_{i=1}^n |I_i|}$,  and we used the derivative-reproducing property of kernels 
\begin{align*}
    \p^{\b p}f(\b x) = \left<f,\p^{\b p,\b 0}k(\cdot,\b x) \right>_{\H_k}.
\end{align*}
Compared to \eqref{eq:k(x,y)} where the kernel values determine the objective, \eqref{eq:kder-finiteD-opt} is determined by the kernel derivatives $\p^{\b p, \b q}k(\b x_i, \b x_j)$.

While kernel techniques are extremely powerful due to their modelling capabilities, this flexibility comes with a price, often they are computationally expensive. 
In order to mitigate this computational bottleneck, several approaches have been proposed in the literature such as the Nystr{\"o}m and sub-sampling methods \cite{williams01using,drineas05nystrom,rudi17falkon}, 
sketching \cite{alaoui15fast,yang17randomized}, or random Fourier features (RFF) \cite{rahimi07random,rahimi08weighted} and their approximate memory-reduced variants and structured extensions \cite{le13fastfood,dai14scalable,bojarski17structured}.

The focus of the current submission is on RFF, arguably the simplest and most influential approximation scheme among these approaches.\footnote{As a recognition of its 
influence, the work \cite{rahimi07random} won the 10-year test-of-time award at NIPS-2017.}
The RFF method constructs a random, low-dimensional, explicit Fourier feature map ($\varphi$) for a continuous, bounded, shift-invariant kernel $k:\R^d \times \R^d \rightarrow \R$ relying on the Bochner's theorem: 
\begin{align*}
    \hat{k}(\b x,\b y) &= \left<\varphi(\b x), \varphi(\b y)\right>, & \varphi&: \R^d \rightarrow \R^m.
\end{align*}
The advantage of such a feature map becomes apparent after applying the parametrization:
\begin{align}
    \hat{f}(\b x) = \left<\b w, \varphi(\b x)\right> \label{eq:fRFF}, \quad \b w \in \R^m.
\end{align}
This parameterization can be considered as an approximate version of the reproducing property 
$ f(\b x) = \left<f,k(\cdot,\b x)\right>_{\H_k}$:
$f\in \H_k$ is changed to $\b w \in \R^m$ and $k(\cdot,\b x)\in \H_k$ to $\varphi(\b x) \in \R^m$.
\eqref{eq:fRFF} allows one to leverage fast solvers 
for kernel machines in the primal [\eqref{eq:J} or \eqref{eq:J2}]. This idea has been applied in a wide range of areas such as 
causal discovery \cite{lopezpaz15towards}, fast function-to-function regression \cite{oliva15fast}, independence testing \cite{zhang17large}, convolution neural networks \cite{cui17kernel}, prediction and filtering in dynamical systems \cite{downey17predictive}, or bandit optimization \cite{li18hyperband}.

Despite the tremendous practical success of RFF-s, its theoretical understanding is quite limited, with only a few optimal guarantees 
\cite{sriperumbudurszabo15optimal,rudi17generalization,sriperumbudur18approximate,li18unified,ullah18streaming,sun18how}.
\begin{itemize}[labelindent=0cm,leftmargin=*,topsep=0cm,partopsep=0cm,parsep=0cm,itemsep=0cm]
 \item Concerning the approximation quality of kernel values, the uniform finite-sample bounds of \cite{rahimi07random,sutherland15error} show that 
      \begin{align*}
	    \big\| k -\widehat{k} \big\|_{L^{\infty}(\K_m \times \K_m)} &: = \sup_{\b x,\b y\in \K_m}\big| k(\b x, \b y) - \widehat{k} (\b x, \b y) \big|\\
								    & =  \O_{p}\left(|\K_m| \sqrt{ \frac{\log m}{m} }\right),     
      \end{align*}
      where $\K_m \subset \R^d$ is  compact, $|\K_m|$ is its diameter, $m$ is the number of RFFs, $\O_p(\cdot)$ means convergence in probability. 
	\cite{sriperumbudurszabo15optimal} recently proved an exponentially tighter finite-sample bound in terms of $|\K_m|$  giving
	\begin{align}
	      \big\| k -\widehat{k} \big\|_{L^{\infty}(\K_m \times \K_m)} = \O_{a.s.}\left(\sqrt{\frac{\log |\K_m|}{m}}\right), \label{eq:k-opt-guarantee}
	\end{align}
	where $\O_{a.s.}(\cdot)$ denotes almost sure convergence. This bound is optimal w.r.t.\ $m$ and $|\K_m|$, as it is known from the characteristic function literature \cite{csorgo83howlong}.
  \item In terms of generalization, \cite{rahimi08weighted} showed that $O(1/\sqrt{n})$ generalization error can be attained using $m=m_n=\O(n)$ RFFs, where $n$ denotes the number of training samples. This bound is somewhat pessimistic, leaving the usefulness of RFFs open. Recently \cite{rudi17generalization} proved  that $\O\left(1 / \sqrt{n}\right)$ generalization 
	  performance is attainable in the context of kernel ridge regression, with $m_n=\o(n)=\O\left(\sqrt{n}\log n\right)$ RFFs. This result settles RFFs in the least-squares setting with Tikhonov regularization. 
  \item \cite{sriperumbudur18approximate} has investigated the computational-statistical trade-offs of RFFs in kernel principal component analysis (KPCA). Their result shows that 
      depending on the eigenvalue decay behavior of the covariance operator associated to the kernel, $m_n=\O(n^{2/3})$ (polynomial decay) or $m_n=\O\left(\sqrt{n}\right)$ (exponential decay) RFFs are sufficient to match 
      the statistical performance of KPCA, where $n$ denotes the number of samples. \cite{ullah18streaming} proved a similar result showing that $m_n=\O\left(\sqrt{n}\log n\right)$ number of RFFs is sufficient for the optimal statistical performance provided that the spectrum of the covariance operator follows an exponential decay, and presented a streaming algorithm for
      KPCA relying on the classical Oja's updates, achieving the same statistical performance.
    \item Results of similar flavour have recently been showed in SVM classification with the 0-1 loss \cite{sun18how}.
\end{itemize}
In contrast to the previous results, the focus of our paper is the investigation of problems involving kernel derivatives [see \eqref{eq:J2} and \eqref{eq:kder-finiteD-opt}]. The idea 
applied in practice is to formally differentiate 
\eqref{eq:fRFF} giving 
\begin{align}
    \widehat{\p^{\b p} f}(\b x) := \p^{\b p} \hat{f}(\b x) = \left<\b w, \p^{\b p}\varphi(\b x)\right>, 
\end{align}
which is then used in the primal [\eqref{eq:J2}], and optimized for $\b w$. From the dual point of view [\eqref{eq:kder-finiteD-opt}], this means that implicitly the kernel derivatives are approximated via RFFs.
The problem we raise in this paper is how accurate these kernel derivative approximations are. 

Our \tb{contribution} is to show that  the same  dependency in terms of $m$ and $|\K|$ can be achieved for kernel derivatives as attained for kernel values (see~\eqref{eq:k-opt-guarantee}). 
To the best of our knowledge, the tightest available
guarantee on kernel derivatives \cite{sriperumbudurszabo15optimal} is 
\begin{align*}
 \big\| \p^{\b{p},\b{q}} k -\widehat{\p^{\b{p},\b{q}} k} \big\|_{L^{\infty}(\K_m \times \K_m)}=\O_{a.s.}\left(|\K_m| \sqrt{\frac{\log m}{m} }\right).
\end{align*}
In this paper, we prove finite sample bounds on the approximation quality of kernel derivatives, which specifically imply that 
\begin{align}
  \hspace*{-0.2cm}\big\| \p^{\b{p},\b{q}} k -\widehat{\p^{\b{p},\b{q}} k} \big\|_{L^{\infty}(\K_m \times \K_m)} &= \O_{a.s.}\left(\sqrt{\frac{\log |\K_m|}{m}} \right). \label{eq:RFFd-opt-dependence}
\end{align}
The possibility of such an exponentially improved dependence in terms of $|\K_m|$ is rather surprising, as in case of kernel derivatives the underlying function classes are no longer uniformly bounded. We circumvent this challenge by applying 
recent tools from unbounded empirical process theory \cite{vandegeer13bernstein}.

Our paper is structured as follows. We formulate our problem in Section~\ref{sec:problem}. The main result on the approximation quality of kernel derivatives is presented in Section~\ref{sec:results}. 
Proofs are provided in Section~\ref{sec:proofs}. 

\section{PROBLEM FORMULATION} \label{sec:problem}
In this section we formulate our problem after introducing a few notations. 

\tb{Notations:} $\N:=\{0,1,2,\ldots\}$, $\N^+:=\N\backslash\{0\}$ and $\R$ denotes the set of natural numbers, positive integers and real numbers respectively. 
For $n\in \N$, $n!$ denotes its factorial. $\Gamma(t)=\int_0^{\infty}x^{t-1}e^{-x}\,\d x$ is the Gamma function ($t>0$); $\Gamma(n+1) = n!$ ($n\in \N$).
Let $n!!$ denote the double factorial of $n \in \N$, that is, the product of all numbers from $n$ to $1$ that have the same parity as $n$; specifically $0!!=1$. If $n$ is a positive odd integer, then 
$n!! = \sqrt{\frac{2^{n+1}}{\pi}} \Gamma\left(\frac{n}{2}+1\right)$. For $n\in\N$, $c_n:=\cos(\frac{\pi n}{2}+\cdot)$ is the $n^{th}$ derivative of the $\cos$ function. For multi-indices $\b{p},\b{q}\in\N^d$, $|\b{p}|=\sum_{j=1}^dp_j$, $\b{v}^{\b{p}}=\prod_{j=1}^d v_j^{p_j}$, and we use 
$\p^{\b{p}}h(\b{x}) := \frac{\partial^{|\b{p}|} h(\b{x})}{\partial_{x_1}^{p_1}\cdots\partial_{x_d}^{p_d}}$, 
$\p^{\b{p},\b{q}}g(\b{x},\b{y}) := \frac{\partial^{|\b{p}|+|\b{q}|} g(\b{x},\b{y})}{\partial^{p_1}_{x_1}\cdots\partial^{p_d}_{x_d} \partial^{q_1}_{y_1}\cdots\partial^{q_d}_{y_d}}$ to denote partial derivatives. $\left<\b a, \b b\right> = \sum_{i=1}^d a_ib_i$ is the inner product between $\b a \in \R^d$ and $\b b \in \R^d$. $\b{a}^T$ is the transpose 
of $\b a\in \R^d$,  $\left\|\b a\right\|_2=\sqrt{\left<\b a, \b a\right>}$ is its Euclidean norm, $\left[\b a_1;\ldots; \b a_M\right] \in \R^{\sum_{m=1}^M d_m}$ is the concatenation of vectors $\b a_m \in \R^{d_m}$.
Let $\K \subset \R^d$ be a Borel set. $\Mp{\K}$ is the set of Borel probability measures on $\K$.  $\Lambda^m = \otimes_{i=1}^m \Lambda$ is the $m$-fold product measure where $\Lambda \in \Mp{\K}$. $L^r(\K)$ is the Banach space of real-valued, $r$-power Lebesgue integrable functions on $\K$ ($1\le r<\infty$). 
$\Lambda f = \int_{\K} f(\bo) \d \Lambda(\bo)$, where $\Lambda \in \Mp{\K}$ and $f\in L^1\left(\K\right)$; specifically for the empirical measure, $\Lambda_m = \frac{1}{m}\sum_{i=1}^m \delta_{\bo_i}$, $\Lambda_m f := \frac{1}{m}\sum_{i=1}^m f(\bo_i)$ where $\bo_{1:m}=(\bo_i)_{i=1}^m \stackrel{\text{i.i.d.}}{\sim}\Lambda$ and $\delta_{\bo}$ is the Dirac measure supported on $\bo \in \K$. 
$\K_{\Delta} = \{\b s_1 - \b s_2: \b s_1, \b s_2 \in \K\}$. For positive sequences $(a_n)_{n\in \N}$, $(b_n)_{n\in \N}$, $a_n = \O(b_n)$ (resp.\ $a_n = \o(b_n)$) means that $\left(\frac{a_n}{b_n}\right)_{n\in \N}$ is bounded (resp.\ $\lim_{n\rightarrow \infty}\frac{a_n}{b_n} = 0$). 
Positive sequences $(a_n)_{n\in \N}$, $(b_n)_{n\in \N}$ are said to be asymptotically equivalent, shortly $a_n \sim b_n$,  if $\lim_{n\rightarrow \infty} \frac{a_n}{b_n}=1$.
$X_n=\O_{p}(r_n)$ (resp.\ $\O_{a.s.}(r_n)$) denotes that $\frac{X_n}{r_n}$ is bounded in probability (resp.\ almost surely).
The diameter of a compact set $A\subset\R^d$ is defined as $|A|:=\sup_{\b x, \b y\in A} \left\|\b x -\b y\right\|_2<\infty$. 
The natural logarithm is denoted by $\ln$.

We continue with the formulation of our task. Let $k:\R^d \times \R^d \rightarrow \R$ be a continuous, bounded, shift-invariant kernel. By the Bochner theorem \cite{rudin90fourier}, it is the Fourier transform of a finite, non-negative Borel measure $\S$:
\begin{align}
  \hspace*{-0.3cm}k(\b{x},\b{y})  &=  \tilde{k}(\b{x}-\b{y}) = \int_{\R^d} e^{\sqrt{-1}\bo^T(\b{x}-\b{y})}\d\S(\bo) \nonumber\\
    & \stackrel{(a)}{=} \int_{\R^d} \cos\left(\bo^T(\b{x}-\b{y})\right)\d\S(\bo) \label{eq:k1} \\ %
    &\stackrel{(b)}{=} \int_{\R^d} \left[ \cos\left(\bo^T\b{x}\right) \cos\left(\bo^T\b{y}\right) + \right. \nonumber \\
    & \hspace{1.2cm} \left. \sin\left(\bo^T\b{x}\right) \sin\left(\bo^T\b{y}\right) \right] \d\S(\bo),\nonumber\\
    &= \int_{\R^d} \left<\phi_{\bo}(\b x), \phi_{\bo}(\b y)\right>_{\R^2}  \d\S(\bo),   \label{eq:k3} 
\end{align}
where $\phi_{\bo}(\b x) = [\cos\left(\bo^T\b{x}\right); \sin\left(\bo^T\b{x}\right)]$.
(a) follows from the real-valued property of $k$, and (b) is a consequence of the trigonometric identity $\cos(\alpha - \beta) = \cos(\alpha) \cos(\beta) + \sin(\alpha) \sin(\beta)$. Without loss of generality, it can be assumed that $\S\in \Mp{\R^d}$ since $\tilde{k}(\b 0) = \S\left(\R^d\right)$ and the normalization $\frac{k(\b{x},\b{y})}{ \tilde{k}(\b 0)}$ yields 
\begin{align*}
 \frac{k(\b{x},\b{y})}{ \tilde{k}(\b 0)} = \int_{\R^d} \cos\left(\bo^T(\b{x}-\b{y})\right) \d\hspace{-0.5cm} \underbrace{\frac{\S(\bo)}{\S\left(\R^d\right)}}_{=:\P(\bo),\,\, \P \in \Mp{\R^d}}\hspace*{-0.4cm}.
\end{align*}

Let $\b{p},\b{q}\in\N^d$. By  differentiating\footnote{By the dominated convergence theorem, the differentiation is valid if 
$\int_{\R^d}|\bo^{\b{p}+\b{q}}|\,\d\Lambda(\bo)<\infty$.} \eqref{eq:k3} one gets
\begin{align}
    \p^{\b{p},\b{q}} k(\b{x},\b{y})   
	      &= \int_{\R^d} \left<\p^{\b p}\phi_{\bo}(\b x), \p^{\b q}\phi_{\bo}(\b y)\right>_{\R^2}  \d\S(\bo). \label{eq:kder}
\end{align}
The resulting expectation can  be approximated by the Monte-Carlo technique using $\bo_{1:m}=(\bo_j)_{j=1}^m\stackrel{\text{i.i.d.}}{\sim}\S$ as
\begin{align}
	\widehat{\p^{\b{p},\b{q}} k}(\b{x},\b{y}) &= \int_{\R^d} \left<\p^{\b p}\phi_{\bo}(\b x), \p^{\b q}\phi_{\bo}(\b y)\right>_{\R^2}  \d\S_m(\bo)\nonumber\\
						    &= \frac{1}{m} \sum_{j=1}^m  \left<\p^{\b p}\phi_{\bo_j}(\b x), \p^{\b q}\phi_{\bo_j}(\b y)\right>_{\R^2} \nonumber\\
					 &= \left< \varphi_{\b{p}}(\b{x}),\varphi_{\b{q}}(\b{y})\right>_{\R^{2m}},\label{eq:RFF-gen}
  \end{align}
where $\S_m= \frac{1}{m} \sum_{j=1}^m \delta_{\bo _j}$, and 
\begin{align}
\varphi_{\b{p}}(\b{x}) &= \frac{1}{\sqrt{m}}  \left( \p^{\b p}\phi_{\bo_j}(\b x) \right)_{j=1}^m  = \p^{\b p} \varphi_{\b 0}(\b x)\in \R^{2m} \label{eq:RFF->der}\\
	     &= \frac{1}{\sqrt{m}}  \left( \bo_j^{\b{p}} \left[ c_{|\b{p}|}\left(\bo_j^T\b{x}\right); c_{3+|\b{p}|}\left(\bo_j^T\b{x}\right)\right] \right)_{j=1}^m.\nonumber
\end{align}
Specifically, if $\b p=\b q=\b 0$ then \eqref{eq:RFF-gen} reduces to the celebrated RFF technique \cite{rahimi07random}:
\begin{align*}
      \widehat{k}(\b x, \b y)  & = \left< \varphi_{\b{0}}(\b{x}),\varphi_{\b{0}}(\b{y})\right>_{\R^{2m}},\\
      \varphi_{\b{0}}(\b{x}) & = \frac{1}{\sqrt{m}} \left( \cos\left(\bo_j^T \b x\right); \sin\left(\bo_j^T \b x\right)\right)_{j=1}^m.
\end{align*}
Our \tb{goal} is to prove that similar to $\b p= \b q = \b 0$ [\eqref{eq:k-opt-guarantee}], fast approximation of kernel derivatives [\eqref{eq:RFFd-opt-dependence}]
is attainable. Alternatively, we establish that the derivative (see $\varphi_\b p$ and \eqref{eq:RFF-gen}-\eqref{eq:RFF->der}) of the RFF feature map ($\varphi_{\b 0}$) is as efficient for kernel derivative approximation as 
$\varphi_{\b 0}$ for kernel value approximation.

\section{MAIN RESULT}\label{sec:results}
In this section we present our main result on the uniform approximation quality of kernel derivatives using RFFs. Its proof is available in Section~\ref{sec:proofs}. 

\begin{theorem}[Uniform guarantee on kernel derivative approximation]\label{thm:uniform}
Suppose that $k: \R^d \times \R^d \rightarrow \R$ is a continuous, bounded and shift-invariant kernel. For $\b p,\b q \in \N^d$, assume 
$C_{\b p, \b q} = \sqrt{ \int_{\R^d} \left|\bo^{\b p + \b q}\right|^2 \left\|\bo\right\|_2^2\d \S(\bo)} / \sigma_{\b p,\b q} < \infty$ and for some constant $K \ge 1$, the following Bernstein condition holds:
\begin{align}
\int_{\R^d} \frac{|\bo^{\b p + \b q}|^n}{\left(\sigma_{\b p,\b q}\right)^n}\d \S(\bo) \le \frac{n!}{2} K^{n-2}, \quad n=2,3,\ldots, \label{eq:Bernstein}
\end{align}
where
$\sigma_{\b p, \b q} = \sqrt{\int_{\R^d}\left|\bo^{\b p + \b q}\right|^2 \d \S(\bo)}$. 
Let 
$L_m = \frac{\sqrt{6}K}{2 \sqrt{m}}$,  $C_1 = 14 \sqrt{6 \ln(2)}+1$,
 $C_2 = 36K[\ln (2)+1]$ and $C_3= 7 \sqrt{6} \left( 1 + \frac{\sqrt{\pi}}{\ln^{\frac{3}{2}}(2)}\right)$.
Then for any $t>0$ and compact set $\K \subset \R^d$,
\begin{align*}
 \S^m\left(\left\{\bo_{1:m}: \big\| \p^{\b{p},\b{q}} k -\widehat{\p^{\b{p},\b{q}} k} \big\|_{L^{\infty}(\K \times \K)} \ge \phantom{\frac{24\sqrt{6}}{\sqrt{m}}} \right. \right.      \\
 \sigma_{\b p, \b q} \left( \frac{C_3 \sqrt{d\ln\left( 16 |\K|C_{\b p, \b q}  + 4 \right)}}{\sqrt{m}} +  \frac{C_1}{\sqrt{m}} +\frac{C_2}{m} +\right.\\ 
 \left. \left.\left.   +  \frac{24\sqrt{6}}{\sqrt{m}}\left[\sqrt{t} + \frac{L_mt}{2}\right]\right)\right\}\right) \le 2e^{-t}.
\end{align*}
\end{theorem}

\tb{Remarks.}
\begin{itemize}[labelindent=0cm,leftmargin=*,topsep=0cm,partopsep=0cm,parsep=0cm,itemsep=0cm]
 \item \tb{Growth of $|\K_m|$}: The theorem proves the same dependence on $m$ and $|\K_m|$ as is known [see \eqref{eq:RFFd-opt-dependence}] for kernel values ($\b p = \b q = \b 0$). The result implies that 
 \begin{align*}
      \big\| \p^{\b{p},\b{q}} k -\widehat{\p^{\b{p},\b{q}} k} \big\|_{L^{\infty}\left(\K_m \times \K_m\right)} \xrightarrow{m\rightarrow \infty} 0 \quad  \text{a.s.}
 \end{align*}
 if $|\K_m| = e^{\o(m)}$. \vspace{0.2cm}
 \item \tb{Requirements for $\b p=\b q=\b 0$:} In this case $\sigma_{\b 0, \b 0} = 1$, 
 \begin{itemize}[labelindent=0cm,leftmargin=*,topsep=0cm,partopsep=0cm,parsep=0cm,itemsep=0cm]
\item $\int_{\R^d} \frac{\left|\bo^{\b 0 + \b 0}\right|^n}{\left(\sigma_{\b 0,\b 0}\right)^n}\d \S(\bo) = 1$, thus \eqref{eq:Bernstein} holds ($K=1$).
 \item The only requirement is the finiteness of $C_{\b 0, \b 0} = \int_{\R^d} \left\|\bo \right\|_2^2 \d \S(\bo)$, which is identical to that imposed in \cite[Theorem~1]{sriperumbudurszabo15optimal} for kernel values.
 \end{itemize} \vspace{0.2cm}
  \item \tb{$L^{\infty}(\K \times \K)$-based $L^r(\K \times \K)$ guarantee}: From the theorem above one can also get (see Section~\ref{sec:proofs}) the following $L^r(\K \times \K)$ guarantee, where $r\in[1,\infty)$. \vspace{0.2cm}
 
 Under the same conditions and notations as in the theorem, for any  $t>0$ 
  \begin{align*}
    \S^m\left(\left\{\bo_{1:m}: \big\| \p^{\b{p},\b{q}} k -\widehat{\p^{\b{p},\b{q}} k} \big\|_{L^r(\K \times \K)} \ge \right.\right.\\
    \sigma_{\b p, \b q} \left[ \frac{\pi^{d/2}|\K|^d}{2^d\Gamma\left(\frac{d}{2}+1\right)} \right]^{\frac{2}{r}}  
    \left( \frac{C_3 \sqrt{2d\ln\left( 16 |\K|C_{\b p, \b q}  + 4 \right)}}{\sqrt{m}} +  \right.\\ 
     \left.\left.\left.  + \frac{C_1}{\sqrt{m}} +\frac{C_2}{m}   + \frac{24\sqrt{6}}{\sqrt{m}}\left[\sqrt{t} + \frac{L_mt}{2}\right] \right) \right\}\right) \le 2e^{-t}.
  \end{align*}
    This shows that     
	$\big\| \p^{\b{p},\b{q}} k -\widehat{\p^{\b{p},\b{q}} k} \big\|_{L^r(\K_m \times \K_m)} = \O_{a.s.}\left(m^{-\frac{1}{2}}|\K_m|^{\frac{2d}{r}} \sqrt{\log |\K_m|}\right)$.
     Consequently, if  
    $|\K_m| \rightarrow \infty$ as $m\rightarrow \infty$ then $\widehat{\p^{\b{p},\b{q}} k}$ is a consistent estimator of $\p^{\b{p},\b{q}} k$ in $L^r(\K_m \times \K_m)$-norm provided that 
    $m^{-\frac{1}{2}}|\K_m|^{\frac{2d}{r}} \sqrt{\log |\K_m|} \xrightarrow{m \rightarrow \infty} 0$. \vspace{0.2cm}
  \item 
        \tb{Bernstein condition with $[\b p; \b q]\ne \b 0$:} Next we illustrate  how the Bernstein condition in \eqref{eq:Bernstein} translates to the efficient estimation of `not too large'-order kernel derivatives in case of the Gaussian kernel. For simplicity 
        let us consider the Gaussian kernel in one dimension ($d=1$); in this case $\S = N\left(0,\sigma^2\right)$ is a normal distribution with mean zero and variance $\sigma^2$. Let 
        $r=p+q\in \N^+$ and denote the l.h.s.\ of \eqref{eq:Bernstein} as  
        \begin{align*}
         A_{r,n}(\S) &= \frac{\int_\R |\omega|^{rn} \d \S (\omega)}{\left[\sqrt{\int_\R \left|\omega\right|^{2r} \d \S(\omega)}\right]^n}.
        \end{align*}
        By the analytical formula for the absolute moments of normal random variables
\begin{align}
A_{r,n}(\S) & = \frac{ \sigma ^{nr}(nr-1)!! \begin{cases} 1 & \text{if } nr \text{ is even}\\ \sqrt{\frac{2}{\pi}}& \text{if } nr \text{ is odd}\end{cases}}{\left[\sigma^{2r}(2r-1)!!\right]^{\frac{n}{2}}} \nonumber
\end{align}
\begin{align}
&= \frac{ (nr-1)!! \begin{cases} 1 & \text{if } nr \text{ is even}\\ \sqrt{\frac{2}{\pi}}& \text{if } nr \text{ is odd}\end{cases}}{\left[(2r-1)!!\right]^{\frac{n}{2}}}. \label{eq:Arn:Normal}
\end{align}
Since $A_{r,n}(\S)$ does not depend on $\sigma$, 
one can assume that $\sigma = \sqrt{\int_\R |\omega|^2 \d \S(\omega)} = 1$ and $\S = N\left(0,1\right)$.
Exploiting the analytical expression obtained for $A_{r,n}(\S)$ one can show (Section~\ref{sec:proofs}) that for  

\begin{itemize}[labelindent=0cm,leftmargin=*,topsep=0cm,partopsep=0cm,parsep=0cm,itemsep=0cm]
 \item $r=1$: \eqref{eq:Bernstein} holds with $K=1$ since $A_{1,n}(\S) \le  \frac{n!}{2}$.
 \item $r=2$: $K=2$ is a suitable choice in \eqref{eq:Bernstein}.
 \item $r=3$ and $r=4$: Asymptotic argument shows that \eqref{eq:Bernstein} can not hold.
\end{itemize} 
It is an interesting open question whether one can relax \eqref{eq:Bernstein} while maintaining similar rates, and what are the trade-offs. \vspace{0.2cm}
  \item \tb{Higher-order derivatives:} In the Gaussian example we saw that \eqref{eq:Bernstein} holds for $r\le 2$, but it is not satisfied for $r>2$. For kernels with
      spectral densities proportional to $e^{-\omega^{2\ell}}$ ($\ell \in \N^+$; the $\ell = 1$ choice reduces to the Gaussian kernel), it turns out that 
      \eqref{eq:Bernstein} is fulfilled with $r\le 2\ell$-order derivatives; for completeness the proof is available in Section~\ref{sec:higher-orderD} (supplement). In other words, kernels with faster decaying spectral densities can guarantee the efficient RFF-based 
      estimation of kernel derivatives, without deterioration in the $|\K|$ and $m$-dependence. \vspace{0.2cm}
  \item \tb{Difficulty:} The fundamental difficulty one has to tackle to arrive at the stated theorem  is as follows. \vspace{0.2cm}
  
	By differentiating \eqref{eq:k1} one gets
	\begin{align*}
	    \p^{\b{p},\b{q}} k(\b{x},\b{y}) \hspace*{-0.07cm} &= \hspace*{-0.09cm} \int_{\R^d} \hspace*{-0.17cm} \bo^{\b{p}}(-\bo)^{\b{q}} c_{|\b{p}+\b{q}|}\left(\bo^T(\b{x}-\b{y})\right) \d\S(\bo).
	\end{align*}
By defining
	      \begin{align}
		  g_{\b{z}}(\bo) &= \bo^{\b{p}}(-\bo)^{\b{q}} c_{|\b{p}+\b{q}|}\left(\bo^T\b{z}\right), \label{eq:gz}
	      \end{align}
the error we would like to control can be rewritten as the supremum of the empirical process
		\begin{eqnarray*}
		  \sup_{\b x,\b y\in \K} \hspace*{-0.09cm}\big|\p^{\b{p},\b{q}} k(\b x, \b y) - \widehat{\p^{\b{p},\b{q}} k} (\b x, \b y) \big|\hspace*{-0.07cm} = \hspace*{-0.09cm}\sup_{\b z\in\K_{\Delta}}\hspace*{-0.09cm} |(\S -\S_m )g_{\b z}|,
		\end{eqnarray*}
		where $\G:=\{g_{\b{z}}: \b{z}\in \K_{\Delta}\}$. For $\b p=\b q=\b 0$ (i.e., the classical RFF-based kernel approximation) 
		\begin{align*}
		    g_{\b z}(\bo) &= \cos\left(\bo^T \b z\right)\quad  (\b z \in \K_\Delta)
		\end{align*}
	      which is a \emph{uniformly bounded} family of functions: 
	      \begin{align*}
		  \sup_{\b z \in \K_\Delta}\left\| g_{\b z}\right\|_{L^{\infty}\left(\R^d\right)} \le 1.
	      \end{align*}
	      This uniform boundedness is the classical assumption 
	      of empirical process theory, which was exploited by \cite{sriperumbudurszabo15optimal} to get the optimal rates. For $\b p, \b q \in \N^d\backslash\{ \b 0\}$, however, the functions $g_{\b z}$ are unbounded and so $\G$ is no longer uniformly bounded in $L^{\infty}\left(\R^d\right)$. Therefore, one has to control unbounded empirical processes for which only few tools are available. \vspace{0.2cm}
	      
	      The key idea of our paper is to apply a recent technique which bounds the supremum as a weighted sum of bracketing entropies of $\G$ at multiple scales. By estimating these bracketing entropies and optimizing the scale the result will follow. This is what we detail in the next section.
	      
\end{itemize}

\section{PROOFS}\label{sec:proofs}

We provide the proofs of the results (main theorem and its consequence, remark on the Bernstein condition for Gaussian kernel) presented in Section~\ref{sec:results}. We start by introducing a few additional notations specific to this section.

\tb{Notations:} The volume of $A\subseteq\R^d$ is defined as $\text{vol}(A)=\int_A 1\, \d\b{x}$.
$\gamma(a,b) = \int_0^b e^{-t} t^{a-1} \d t$ is the incomplete Gamma function ($a>0$, $b\ge 0$) that satisfies
$\gamma(a+1,b) = a \gamma(a,b) - b^ae^{-b}$ and $\gamma\left(\frac{1}{2},b\right) = \sqrt{\pi}\text{erf}\left(\sqrt{b}\right)$, where $\text{erf}(b) = \frac{2}{\sqrt{\pi}} \int_{0}^b e^{-t^2} \d t$ is the error function ($b\ge 0$).
Let $(\F,\rho)$ be a metric space. The $r$-covering number of $\F$ is defined as the size of the smallest $r$-net, i.e., 
  $N(r,\F,\rho)  =
  \inf\left\{\ell\ge 1: \exists\, (f_j)_{j=1}^\ell \text{ s.t. } \F \subseteq \cup_{j=1}^\ell B_{\rho}(f_j,r)\right\}$,
where $B_{\rho}(s,r) = \{f\in \F: \rho(f,s)\le r\}$ is the closed ball with center $s\in \F$ and radius $r$. 
For a set of real-valued functions $\F$ and $r>0$, the cardinality of the minimal $r$-bracketing of $\F$ is defined as $N_{[\hspace{0.06cm}]}(r,\F,\rho) = \inf\{n \ge 1: \exists\, 
\{\left(f_{j,L},f_{j,U}\right)\}_{j=1}^n, f_{j,L}, f_{j,U}\in \F\,\, (\forall j)$   such that $\rho\left(f_{j_L},f_{j,U}\right)\le r$ and $\forall f\in \F\,\, \exists j\,\, f_{j,L} \le f \le f_{j,U}\}$. 

The \tb{proof of the main theorem} is structured as follows.
\begin{compactenum}
  \item First, we rescale and reformulate the approximation error as the suprema of unbounded empirical processes, for which bounds in terms of bracketing entropies at multiple scales can be obtained.\vspace{.5mm}
  \item Then, we bound the bracketing entropies via Lipschitz continuity.\vspace{.5mm}
  \item Finally, the scale is optimized.\vspace{2mm}
\end{compactenum}

\tb{Step 1.}
	It follows from \eqref{eq:gz} that, 
	\begin{align*}
	  \left\|g_{\b z}\right\| &:= \left\|g_{\b z}\right\|_{L^2(\R^d,\Lambda)} = \sqrt{\S g_{\b z}^2}\nonumber\\
& \le  \underbrace{\sqrt{\int_{\R^d}\left|\bo^{\b p + \b q}\right|^2 \d \S(\bo)}}_{=: \sigma_{\b p, \b q}}.
	\end{align*}
	Define $f_{\b{z}}(\bo):= \frac{g_{\b{z}}(\bo)}{\sigma_{\b p, \b q}}$ so that
	\begin{align}
	     \left\| f_{\b z} \right\| \le 1 \quad  \forall \b{z}\in \K_{\Delta} \hspace{0.1cm} \Rightarrow \hspace{0.1cm} \sup_{f\in \F} \left\|f\right\| \le 1, \label{eq:normalized-gs}
	\end{align}
	where $\F:=\{f_{\b{z}}: \b{z}\in \K_{\Delta}\}$. The target quantity can be rewritten in supremum of empirical process form as 
	\begin{eqnarray*}
	  \lefteqn{\sup_{\b x,\b y\in \K}\big|\p^{\b{p},\b{q}} k(\b x, \b y) - \widehat{\p^{\b{p},\b{q}} k} (\b x, \b y) \big| = \sup_{\b z\in\K_{\Delta}}|\S g_\b z-\S_m g_{\b z}|}\\
	   &&\hspace*{0.6cm} = \sigma_{\b p, \b q} \sup_{f\in\F}\left|(\S -\S_m)f\right| =: \sigma_{\b p, \b q} \left\|\S-\S_m\right\|_{\F}.
	\end{eqnarray*}
By the Bernstein condition [\eqref{eq:Bernstein}] the following uniform bound holds:
	\begin{align}
	    \sup_{f_{\b{z}}: \b z\in \K_{\Delta}} \Lambda |f_{\b z}|^n &\le \int_{\R^d} \frac{|\bo^{\b p + \b q}|^n}{\left(\sigma_{\b p,\b q}\right)^n}\d \S(\bo) \nonumber\\ 
								       &\le \frac{n!}{2} K^{n-2}\quad  (n=2,3,\ldots).\label{eq:Bernstein-on-gs}
	\end{align}
The uniform $L^2(\S)$ boundedness of $\F$ [\eqref{eq:normalized-gs}] with its Bernstein property [\eqref{eq:Bernstein-on-gs}] imply by \cite[Theorem~8]{vandegeer13bernstein}  that for all $t>0$ and for 
	all scale $S\in \N$
	\begin{align}
	  \S^m\left(\left\{\bo_{1:m}: \sup_{f\in \F} \left|\sqrt{m}( \S - \S_m) f\right| \ge \min_S E_S \right.\right.\label{eq:concentration}\\
	  \left.\left. +\frac{36 K}{\sqrt{m}} + 24\sqrt{6}\left[\sqrt{t} + \frac{L_mt}{2}\right]\right\}\right) \le 2e^{-t}, \nonumber
	\end{align}
	where 
	\begin{align*}
	    E_S &:= 2^{-S} \sqrt{m} + 14 \sum_{s=0}^S 2^{-s} \sqrt{6 H_s} + \frac{36 K H_0}{\sqrt{m}},\\
	      L_m &:=  \frac{\sqrt{6}K}{2 \sqrt{m}}, \quad H_s := \ln(N_s+1),\\
	      N_s &:= N_{[\hspace{0.06cm}]}(2^{-s},\F,\left\|\cdot\right\|),\\
	      H_0 &= \ln(N_0 + 1),
	\end{align*}
	and $N_0$ is the  cardinality of the minimal generalized bracketing set of $\F$. Formally, 
	$N_0 = N_0(K):= \inf\{n\ge 1:\, \exists f_{j,L}, f_{j,U} \in \F\, (j=1,\ldots,n),\, \Lambda \left|f_{j,L}-f_{j,U}\right|^n \le \frac{n!}{2}(2K)^{n-2}\, (n=2,3,\ldots)$, and for $\forall f\in\F,
	\,\, \exists j\in\{1,\ldots,n\}$ such that $f_{j,L}\le f \le f_{j,U}\}$.
	
\tb{Step 2.} We continue the proof by bounding the entropies $H_0$ and $H_s$ $(s\ge 1)$ in \eqref{eq:concentration}. Using \eqref{eq:Bernstein}
	for the envelope function $F:= \sup_{f\in \F} |f|$, we get
	\begin{align*}
	 \Lambda \left(F^n\right) &= \Lambda\left( \Big[\sup_{f\in \F}|f|\Big]^n\right) = \Lambda\left(\sup_{f\in \F} |f|^n\right) \\
				   & \le \int_{\R^d} \frac{|\bo^{\b p + \b q}|^n}{\left(\sigma_{\b p,\b q}\right)^n}\d \S(\bo) \le \frac{n!}{2} K^{n-2},\,\, n=2,3,\ldots
	\end{align*}
	Hence $F$ also  satisfies the weaker Bernstein condition: $\Lambda \left(F^n\right) \le \frac{n!}{2} (2K)^{n-2}$ $(n=2,3,\ldots)$. Consequently, one can choose $N_0 = 1$ 
	\cite[remark after Definition~8]{vandegeer13bernstein}, and $H_0 = \ln(N_0+1) = \ln (2)$.
	
	Next we bound $H_s$ ($s\ge 1$). The $\F$ function class is Lipschitz continuous in the parameters ($f_{\b z_1}, f_{\b z_2}\in \F$): 
	\begin{eqnarray*}
	 \lefteqn{|f_{\b z_1}(\bo) - f_{\b z_2}(\bo)| } \\
		    && \hspace*{-0.7cm} = \frac{\left| \bo^{\b{p}}(-\bo)^{\b{q}} c_{|\b{p}+\b{q}|}\left(\bo^T\b{z}_1\right) - \bo^{\b{p}}(-\bo)^{\b{q}} c_{|\b{p}+\b{q}|}\left(\bo^T\b{z}_2\right) \right|}{\sigma_{\b p,\b q}}\\
					     && \hspace*{-0.7cm} = \frac{\left|\bo^{\b p + \b q}\right| \left|c_{|\b{p}+\b{q}|}\left(\bo^T\b{z}_1\right) - c_{|\b{p}+\b{q}|}\left(\bo^T\b{z}_2\right)\right|}{\sigma_{\b p,\b q}}\\
					     &&\hspace*{-0.7cm} \stackrel{(a)}{\le} \frac{\left|\bo^{\b p + \b q}\right|}{\sigma_{\b p,\b q}} \left|\b \bo^T (\b z_1 - \b z_2)\right|
					     \stackrel{(b)}{\le} \underbrace{\frac{\left|\bo^{\b p + \b q}\right|}{\sigma_{\b p,\b q}}\left\|\bo \right\|_2}_{=:G(\bo)} \left\|\b z_1- \b z_2\right\|_2,
	\end{eqnarray*}
where we used the Lipschitz property of $u \mapsto c_{|\b p + \b q|}(u)$ (with Lipschitz constant $1$) in (a) and the Cauchy-Bunyakovskii-Schwarz inequality in (b). Thus, by \cite[Theorem~2.7.11, page~164]{vandervaart96weak} for any $\delta >0$,
	\begin{align}
	 N_{[\hspace{0.06cm}]}(\delta, \F, \left\|\cdot\right\|) &\le N\left(\frac{\delta}{2 \left\|G\right\|},\K_{\Delta},\left\| \cdot \right\|_2\right), \label{eq:Nbracket}
	\end{align}
	where 
	\begin{align*}
	    \left\|G\right\| &= \sqrt{\int_{\R^d} G^2(\bo) \d \S(\bo)}\\
			    &= \sqrt{ \int_{\R^d} \frac{\left|\bo^{\b p + \b q}\right|^2}{\sigma^2_{\b p,\b q}} \left\|\bo\right\|_2^2\d \S(\bo)} =: C_{\b p, \b q}.
	\end{align*}
	From Lemma 2.5 in \cite{geer09empirical} it follows that
	\begin{align*}
	    N\left(r,M,\left\|\cdot\right\|_2\right) \le \left(\frac{2 |M|}{r}+1\right)^d, \quad \forall r>0
	\end{align*}
	for any compact $M\subset \R^d$. Choosing $M=\K_{\Delta}$, $\delta = 2^{-s}$ and noting that $|\K_{\Delta}| \le 2| \K|$, one can bound the l.h.s. in \eqref{eq:Nbracket} as 
	\begin{align*}
	    N_s &= N_{[\hspace{0.06cm}]}\left(2^{-s}, \F, \left\|\cdot\right\|\right) \le N\left(\frac{1}{2^{s+1}C_{\b p, \b q}},\K_{\Delta},\left\| \cdot \right\|_2\right)\\
		&\le \Big(\underbrace{2^{s+3} |\K| C_{\b p, \b q} + 1}_{\le 2^s \tilde{K}_{|S|}}\Big)^{d},
	\end{align*}
	where $\tilde{K}_{|S|}=8 |\K| C_{\b p, \b q} + 1$. 
	Thus for any $s\ge 1$, 
	\begin{align*}
	    H_s &= \ln(N_s+1) \le d \ln\big(\underbrace{2^s \tilde{K}_{|S|}+1}_{\le 2^s(\tilde{K}_{|\K|} + 1)}\big)\\
	      &\le  d\left[s\ln (2)+\ln\left(\tilde{K}_{|\K|}+1\right)\right]\\
	      & \le s \underbrace{d\left[\ln (2)+\ln\left(\tilde{K}_{|\K|}+1\right)\right]}_{d\ln\left(2 \tilde{K}_{|\K|} + 2\right) = :K_{|\K|}}.
	\end{align*}
	Hence, 
	\begin{align}
	      E_S &\le \underbrace{14 \sum_{s=1}^S 2^{-s} \sqrt{6 s K_{|\K|}}}_{14 \sqrt{6 K_{|\K|}}\sum_{s=1}^S 2^{-s}\sqrt{s}} +2^{-S} \sqrt{m}\nonumber\\
	      & \quad\quad\quad  + 14 \sqrt{6 \ln (2)} + \frac{36 K \ln (2)}{\sqrt{m}}.\label{eq:E_S}
	\end{align}

\tb{Step 3.} By \eqref{eq:E_S}, to control $E_S$ as a function of the scale $S$, we study the behaviour of $h(t)= 2^{-t}\sqrt{t}$. It is easy to verify that $h$ is monotonically decreasing on $\left[\frac{1}{2\ln (2)},\infty\right)$ as its derivative
      \begin{align*}
	  h'(t) &= \frac{\frac{1}{2}t^{-\frac{1}{2}}2^t - \sqrt{t} 2^{t} \ln (2)}{2^{2t}}\le 0
\end{align*}
on $\left[\frac{1}{2\ln (2)},\infty\right)$.
Using this monotonicity, one gets $h(s) \le \int_{s-1}^{s}h(x) \d x$ for any $s$ such that $\frac{1}{2\ln (2)} \le s-1 \Leftrightarrow \frac{1}{2\ln (2)}+1 \le s$, specifically for all $2 \le s$ 
      since $\frac{1}{2 \ln (2)}< 1$. Hence, applying change of variables ($2^{-x} = e^{-t}$, i.e.\ $x=\frac{t}{\ln (2)}$) we arrive at 
      \begin{eqnarray*}
	\lefteqn{\sum_{s=1}^S 2^{-s}\sqrt{s} = \underbrace{h(1)}_{\frac{1}{2}} + \sum_{s=2}^S \underbrace{h(s)}_{\le \int_{s-1}^s h(x) \d x} \hspace*{-0.04cm} \le  \frac{1}{2} + \int_{1}^S h(x)  \d x} \\
	&&\hspace*{-.3cm}=  \frac{1}{2} +  \frac{1}{\ln^{\frac{3}{2}}(2)}\int_{\ln (2)}^{S\ln (2)} e^{-t} \sqrt{t} \d t\\
	&&\hspace*{-.3cm}\le  \frac{1}{2} + \frac{1}{\ln^{\frac{3}{2}}(2)}\int_{0}^{S\ln (2)} e^{-t} \sqrt{t} \d t\\ 
	&&\hspace*{-.3cm}=  \frac{1}{2} +  \frac{1}{\ln^{\frac{3}{2}}(2)}\left[\frac{\sqrt{\pi}}{2} \text{erf}(\sqrt{S\ln (2)})-2^{-S}\sqrt{S \ln (2)}\right].\nonumber
      \end{eqnarray*}
      Plugging this estimate in \eqref{eq:E_S}  results in
      \begin{align*}
	E_S & \le  \frac{\sqrt{m}}{2^S} + 14 \sqrt{6 \ln (2)}+ \frac{36 K \ln (2)}{\sqrt{m}} + 14 \sqrt{6 K_{|\K|}} \times \\
	& \hspace*{-0.5cm}\, \times \left(\frac{1}{2}+ \hspace*{-0.05cm} \frac{1}{\ln^{\frac{3}{2}}(2)} \left[ \frac{\sqrt{\pi}}{2} \text{erf}\left(\sqrt{S\ln (2)}\right) -  \frac{\sqrt{S\ln (2)}}{2^S}\right]\right)\\
	& \le  \frac{\sqrt{m}}{2^S} +  14 \sqrt{6}  \sqrt{K_{|\K|}} \times\left( \frac{1}{2}+ \frac{1}{\ln^{\frac{3}{2}}(2)} \frac{\sqrt{\pi}}{2} \right) \\
& \quad + C_1 +\frac{C_2}{\sqrt{m}}\nonumber\\
	& \le \frac{\sqrt{m}}{2^S} + 14 \sqrt{6} \sqrt{d\ln\left( 16 |\K|C_{\b p, \b q}  + 4 \right)} \times \\
	& \quad \times \left( \frac{1}{2}+ \frac{1}{\ln^{\frac{3}{2}}(2)} \frac{\sqrt{\pi}}{2} \right) +  C_1 +\frac{C_2}{\sqrt{m}} =:(*),
      \end{align*}
where we used the fact that $\text{erf}(b) \le 1$ for any $b\ge 0$, $2^{-S} \sqrt{S}\ge 0$, $C_1 = 14 \sqrt{6 \ln (2)}$, $C_2 = 36K\ln (2)$ and $K_{|\K|}=d\ln\left(2\tilde{K}_{|\K|}+2\right) = d\ln\left( 16 |\K| C_{\b p, \b q} + 4 \right)$. Let us choose the scale $S$ such that $2^{-S} \sqrt{m} \le 1$, i.e.\ $\frac{\ln (m)}{2\ln (2)} \le S$. In this case, by defining $C_3 = 7 \sqrt{6} \left( 1 + \frac{\sqrt{\pi}}{\ln^{\frac{3}{2}}(2)}  \right)$, we have
      \begin{align*}
	(*) &= 1 + C_3 \sqrt{d\ln\left( 16 |\K|C_{\b p, \b q}  + 4 \right)}  + C_1 +\frac{C_2}{\sqrt{m}}.
      \end{align*}
      Combining this result with \eqref{eq:concentration}, we obtain
      \begin{eqnarray*}
      \lefteqn{\S^m\left(\left\{\bo_{1:m}: \left\| \S - \S_m \right\|_{\F} \ge \frac{C_3 \sqrt{d\ln\left( 16 |\K|C_{\b p, \b q}  + 4 \right)}}{\sqrt{m}} \right.\right.} \\
      && \hspace{-1.1cm}\quad  + \hspace*{-0.06cm}\left.\left. \frac{C_1 +1}{\sqrt{m}} \hspace*{-0.06cm}+\hspace*{-0.06cm}\frac{C_2+36K}{m} \hspace*{-0.06cm}+ \hspace*{-0.06cm} \frac{24\sqrt{6}}{\sqrt{m}}\hspace*{-0.06cm}\left[\sqrt{t} \hspace*{-0.04cm} + \hspace*{-0.06cm} \frac{L_mt}{2}\right]\right\}\right) \le 2e^{-t}.
      \end{eqnarray*}
      By redefining $C_1$ and $C_2$ as $C_1 = 14 \sqrt{6 \ln (2)}+1$, $C_2 = 36K[\ln (2)+1]$ and taking into account the $\sigma_{\b p, \b q}$ normalization, the claimed result follows.\qed

The \tb{proof of the consequence}  is as follows. Let $r\in[1,\infty)$ be fixed. Then
\begin{eqnarray*}
  \lefteqn{\big\| \p^{\b{p},\b{q}} k -\widehat{\p^{\b{p},\b{q}} k} \big\|_{L^r(\K \times \K)} = }\\
  &&= \left(\int_\K \int_\K \left| \p^{\b{p},\b{q}} k(\b x, \b y) - \widehat{\p^{\b{p},\b{q}} k}(\b x,\b y) \right|^{r} \d \b x \d \b y\right)^{\frac{1}{r}}\\
  &&\le \left(\int_\K \int_\K \big\| \p^{\b{p},\b{q}} k -\widehat{\p^{\b{p},\b{q}} k} \big\|_{L^{\infty}(\K \times \K)} ^{r} \d \b x \d \b y\right)^{\frac{1}{r}}\\
  &&= \left[ \big\| \p^{\b{p},\b{q}} k -\widehat{\p^{\b{p},\b{q}} k} \big\|_{L^{\infty}(\K \times \K)}^{r} \vol^2(\K)\right]^{\frac{1}{r}}\\
  &&= \big\| \p^{\b{p},\b{q}} k -\widehat{\p^{\b{p},\b{q}} k} \big\|_{L^{\infty}(\K \times \K)} \vol^{\frac{2}{r}}(\K).
\end{eqnarray*}
Using the fact (which follows from \cite[Corollary 2.55]{folland99real}) that $\vol(\K) \le \frac{\pi^{d/2}|\K|^d}{2^d\Gamma\left(\frac{d}{2}+1\right)}$, we obtain
\begin{eqnarray*}
 \lefteqn{\big\| \p^{\b{p},\b{q}} k -\widehat{\p^{\b{p},\b{q}} k} \big\|_{L^r(\K\times \K)} \le}\\
 && \le \big\| \p^{\b{p},\b{q}} k -\widehat{\p^{\b{p},\b{q}} k} \big\|_{L^{\infty}(\K\times \K)}  \left[ \frac{\pi^{d/2}|\K|^d}{2^d\Gamma\left(\frac{d}{2}+1\right)} \right]^{\frac{2}{r}}.
\end{eqnarray*}
Hence the main theorem implies the claimed $L^r(\K \times \K)$ bound.\qed

The result on the \tb{Bernstein condition} for the Gaussian kernel can be obtained as follows. Recall that the goal is to check \eqref{eq:Bernstein} and we apply the expression for $A_{r,n}(\S)$ given in \eqref{eq:Arn:Normal}.
\begin{itemize}[labelindent=0cm,leftmargin=*,topsep=0cm,partopsep=0cm,parsep=0cm,itemsep=0cm]
\item  For $r=1$: 
    \begin{align*}
	A_{1,n}(\S) &= \int_\R |\omega|^n\d \S(\omega)\\
	&= (n-1)!! \begin{cases} 1 & \text{if } n \text{ is even}\\ \sqrt{\frac{2}{\pi}}& \text{if } n \text{ is odd}\end{cases}\nonumber\\ 
&\le (n-1)!!  \le (n-1)! \le \frac{n!}{2},
    \end{align*}
    where the last inequality is equivalent to $2\le n$. Hence, \eqref{eq:Bernstein} is satisfied with $K=1$.
\item For $r=2$: In this case $nr$ is even and $A_{2,n}(\S) = \frac{(2n-1)!!}{3^{\frac{n}{2}}}$ by \eqref{eq:Arn:Normal}. For \eqref{eq:Bernstein}, it is enough ($K^{n-2}\le K^n$) that for some $K \ge 1$ and for $n=2,3,\ldots$
	\begin{align}
	A_{2,n}(\S) \le \frac{n!}{2}K^n &\Leftrightarrow \underbrace{(2n-1)!!}_{\frac{2^n}{\sqrt{\pi}}\Gamma\left(n + \frac{1}{2}\right)} \le \underbrace{n!}_{\Gamma(n+1)} \frac{1}{2} \left(\sqrt{3}K\right)^n \nonumber\\ 
	&\stackrel{(a)}{\Leftarrow} \frac{2^n}{\sqrt{\pi}} \le \frac{1}{2} \left(\sqrt{3}K\right)^n \nonumber\\
	&\Leftrightarrow \frac{2}{\sqrt{\pi}} \le \left(\frac{\sqrt{3}K}{2}\right)^n. \label{eq:r=2:b}
	\end{align}
	In (a) we used that $\Gamma\left(n + \frac{1}{2}\right) \le \Gamma(n+1)$ for $n\ge 2$. \eqref{eq:r=2:b} holds e.g.\ with $K=2$ since $1 < \frac{2}{\sqrt{\pi}} < \sqrt{3}$.
 \item For $r=3$:  Let us restrict $n$ to even numbers ($n=2\ell$, $\ell\in \N^+$) in \eqref{eq:Bernstein}. By \eqref{eq:Arn:Normal},
	  $A_{3,n}(\Lambda) = \frac{(3n-1)!!}{(5!!)^{\frac{n}{2}}}$, and \eqref{eq:Bernstein} can be written as
	  \begin{align*}
	    \underbrace{\frac{(6\ell-1 )!!}{15^{\ell}}}_{\frac{2^{3\ell}}{\sqrt{\pi}} \Gamma\left(3\ell + \frac{1}{2}\right) \frac{1}{15^{\ell}}} \le \frac{(2\ell)!}{2} K^{2 \ell -2}, \quad \forall \ell\in \N^{+}.
	  \end{align*}
	  Using the bound, $\Gamma\left(3\ell + \frac{1}{2}\right) \ge  \Gamma\left(3\ell\right) = (3\ell - 1)!$, we have that 
	  \begin{align*}
	      \left(\frac{8}{15}\right)^{\ell} \frac{1}{\sqrt{\pi}} (3\ell - 1)! \le \frac{(2\ell)!}{2} K^{2 \ell -2}, \quad \forall \ell\in \N^{+}
	  \end{align*}
	  should also hold. By the Stirling's formula $u! \sim \sqrt{2\pi u} \left(\frac{u}{e}\right)^u$, we have
	  \begin{align*}
	  \left(\hspace*{-0.02cm}\frac{8}{15}\hspace*{-0.02cm}\right)^{\ell} \hspace*{-0.11cm} \frac{\sqrt{2\pi (3\ell -1)}}{\sqrt{\pi}} \hspace*{-0.03cm} \left(\hspace*{-0.03cm}\frac{3\ell-1}{e}\hspace*{-0.03cm}\right)^{3\ell-1}\hspace*{-0.25cm} \le\hspace*{-0.03cm} \frac{\sqrt{2\pi (2\ell)}}{2} \hspace*{-0.02cm} \left(\hspace*{-0.02cm}\frac{2\ell}{e}\hspace*{-0.02cm}\right)^{2\ell}
	  \end{align*}
	  as $\ell \rightarrow \infty$. Taking $\ln(\cdot)$ yields
	  \begin{align*}
	  \ln(\text{l.h.s.}) & = \ell \ln\left(\frac{8}{15}\right) + \ln\left(\sqrt{2\pi (3\ell -1)}\right) \\
	  &  \quad + (3\ell - 1) \left[\ln(3\ell -1) - 1\right]+\ln\left(1/\sqrt{\pi}\right) \\
	      & \sim  (3\ell - 1)\ln(3\ell -1),\\
	  \ln(\text{r.h.s.}) & = \ln \left(\sqrt{2\pi (2\ell)} \right)  - \ln (2) + 2\ell [\ln(2 \ell) - 1]\\
	  &\sim 2\ell \ln(2 \ell).
	  \end{align*}
	  Since $\ln(\text{l.h.s.})$ is asymptotically larger than $\ln(\text{r.h.s.})$, \eqref{eq:Bernstein} can not hold.
  \item For $r=4$: $nr$ is even, $A_{4,n}(\S) = \frac{(4n-1)!!}{[7!!]^{\frac{n}{2}}}$ by \eqref{eq:Arn:Normal}, and \eqref{eq:Bernstein}  is equivalent to  
  	\begin{align*}
	       \underbrace{(4n-1)!!}_{\frac{2^{2n}}{\sqrt{\pi}} \Gamma\left(2n+\frac{1}{2}\right)} \le \frac{n!}{2}K^{n-2}  \left(\sqrt{7\times 5 \times 3 }\right)^n.
	\end{align*}
	By using the $\Gamma(z+1) = z \Gamma(z)$ recursion, we obtain
	\begin{align*}
	    \Gamma\left(2n + \frac{1}{2}\right) \hspace*{-0.04cm}&= \hspace*{-0.04cm}\underbrace{\underbrace{\left(2n-\frac{1}{2}\right)}_{1.} \underbrace{\left(2n-\frac{3}{2}\right)}_{2.} \cdots \underbrace{\left(n+\frac{3}{2}\right)}_{n-1.}}_{\ge (n-1)^{n-1}} \times\\
	    & \quad \times \Gamma\left(n+\frac{3}{2}\right) \le \underbrace{n!}_{\Gamma(n+1)}\hspace{0.2cm} \underbrace{K^{n-2}}_{K^{-1}K^{n-1}}.
	\end{align*}
	Since $\Gamma\left(n+\frac{3}{2}\right) >  \Gamma(n+1)$ for all $n\in \N^+$ and $f(n)=n^n$ grows faster than $g(n)=K^n$ for any fixed $K$, \eqref{eq:Bernstein} 
	can not be satisfied for all $n\ge 2$.
\end{itemize}

\subsubsection*{Acknowledgements}
This work was started and partially carried out while ZSz was visiting BKS at the Department of Statistics, Pennsylvania State University; ZSz thanks for their generous support. BKS is supported by NSF-DMS-1713011.

\bibliography{./BIB/sample_paper}
\bibliographystyle{plain}

\clearpage
\appendix

\begin{center}
 {\Large\tb{Supplement}}
\end{center}
In Section~\ref{sec:higher-orderD} we prove our remark on the validity of the Bernstein condition for higher-order derivatives in the case of kernels with faster spectral decay. 
The result extends the example of Gaussian kernels detailed in the main part of the paper.

\section{BERNSTEIN CONDITION FOR HIGHER-ORDER DERIVATIVES} \label{sec:higher-orderD} We prove that in the case of kernels with spectral density decaying as $f_\S(\omega)\propto e^{-\omega^{2\ell}}$ ($\ell \in \N^+$), the Bernstein 
condition \eqref{eq:Bernstein}  holds for $r\le 2\ell$-order derivatives. This example extends the case of Gaussian kernels where $\ell = 1$ and $r\le 2$. Let $\ell \in \N^+$ and the spectral measure associated with kernel $k$ be absolutely continuous w.r.t.\ the Lebesgue measure with density 
\begin{align*}
  f_{\S}(\omega) &=  c_\ell e^{-\omega^{2\ell}}
\end{align*}
for some $c_\ell>0$. $f_{\S}$ is positive and we determine $c_\ell$ as:
\begin{align*}
      1 &= \int_{\R} f_\S(\omega)\d \omega = \int_{\R} c_\ell e^{-\omega^{2\ell}} \d \omega = 2 c_\ell \int_0^{\infty} e^{-\omega^{2\ell}} \d \omega\\
      &= 
      \frac{c_\ell}{\ell} \int_0^\infty e^{-y}y^{\frac{1}{2\ell}-1} \d y = \frac{c_\ell}{\ell} \Gamma\left(\frac{1}{2\ell}\right) \quad \Rightarrow \\
      c_\ell &= \frac{\ell}{\Gamma\left(\frac{1}{2\ell}\right)}
\end{align*}
where we used $y = \omega^{2 \ell}$, $\omega = y^{\frac{1}{2\ell}}$, $\d \omega = \frac{1}{2 \ell} y^{\frac{1}{2\ell}-1} \d y$ and the pdf of the Gamma distribution ($b=1$, $a=\frac{1}{2\ell}$)
$g(y;a,b) =\frac{b^a}{\Gamma(a)} y^{a-1}e^{-by}$, ($y>0$, $a>0$, $b>0$) from which it follows that 
\begin{align}
  \int_0^{\infty} y^{a-1}e^{-by} \d y&= \frac{\Gamma(a)}{b^a}. \label{eq:gamma}
\end{align}
Consequently, one obtains
\begin{align*}
 A_{r,n} &= A_{r,n}(\S) = \frac{\int_\R |\omega|^{rn} \d \S (\omega)}{\left[\sqrt{\int_\R \left|\omega\right|^{2r} \d \S(\omega)}\right]^n} 
	     = \frac{\frac{\Gamma\left(\frac{rn+1}{2\ell}\right)}{\Gamma\left(\frac{1}{2\ell}\right)}}{ \left[\frac{\Gamma\left(\frac{2r+1}{2 \ell}\right)}{\Gamma\left(\frac{1}{2 \ell}\right)}\right]^{\frac{n}{2}}}
\end{align*}
by using \eqref{eq:gamma} with $b=1$, $a=\frac{r+1}{2 \ell}$ and the value of $c_\ell$:
\begin{eqnarray*}
      \lefteqn{\int_{\R} |\omega|^r \d \S(\omega) = \int_{\R} |\omega|^r c_{\ell} e^{-\omega^{2\ell}} \d \omega}\\
       && = 2 c_{\ell} \int_0^{\infty} \omega^r  e^{-\omega^{2\ell}} \d \omega = \frac{c_{\ell}}{\ell} \int_0^\infty e^{-y}y^{\frac{r}{2\ell}}y^{\frac{1}{2\ell}-1} \d y\\
       &&=  \frac{c_{\ell}}{\ell} \Gamma\left(\frac{r+1}{2\ell}\right)
      = \frac{\Gamma\left(\frac{r+1}{2\ell}\right)}{\Gamma\left(\frac{1}{2\ell}\right)}.
\end{eqnarray*}
Next we assume that $r\le 2\ell$ is fixed and apply induction to prove \eqref{eq:Bernstein}. 

\begin{itemize}[labelindent=0cm,leftmargin=*,topsep=0cm,partopsep=0cm,parsep=0cm,itemsep=0cm]
    \item For $n=2$, by definition $A_{r,2} = 1$ ($\forall r\in\N^+$). 
    \item The induction argument is as follows.  By the inductive assumption it is sufficient to 
	show the existence of $K_r\ge 1$ such that 
      \begin{align}
	  B_{r,n} &:= \frac{A_{r,n+1}}{A_{r,n}} \le (n+1) K_r \label{eq:B-bound}
      \end{align}
      since $A_{r,n}\le \frac{n!}{2}K_r^{n-2}$ and $\frac{A_{r,n+1}}{A_{r,n}} \le (n+1) K_r$ imply $A_{r,n+1}\le \frac{(n+1)!}{2}K_r^{n+1-2}$. By defining $c_r := \frac{\Gamma\left(\frac{2r+1}{2 \ell}\right)}{\Gamma\left(\frac{1}{2\ell}\right)}$, we obtain
      \begin{align*}
	  B_{r,n} &= \frac{\Gamma\left(\frac{r(n+1)+1}{2 \ell}\right)}{\Gamma\left(\frac{1}{2 \ell}\right) (c_r)^{\frac{n+1}{2}}} 
	  \frac{\Gamma\left(\frac{1}{2 \ell}\right) (c_r)^{\frac{n}{2}}}{\Gamma\left(\frac{rn+1}{2 \ell}\right)}
	  = \frac{\Gamma\left(\frac{r(n+1)+1}{2\ell}\right)}{\sqrt{c_r}\,\,\Gamma\left(\frac{rn+1}{2\ell}\right)}\\
	  &= \frac{\Gamma\left(\frac{rn+1}{2\ell} + \frac{r}{2\ell}\right)}{\sqrt{c_r}\,\,\Gamma\left(\frac{rn+1}{2\ell}\right)}
	  \stackrel{(a)}{\le} D_{r,n} \frac{\Gamma\left(\frac{rn+1}{2\ell} + \frac{2\ell}{2\ell}\right)}{\sqrt{c_r}\,\,\Gamma\left(\frac{rn+1}{2\ell}\right)}\\
	  &\stackrel{(b)}{=} D_{r,n} \frac{1}{\sqrt{c_r}}\frac{rn +1}{2 \ell} \stackrel{(c)}{\le } D_{r,n} \frac{1}{\sqrt{c_r}}\underbrace{\frac{2\ell n +1}{2 \ell}}_{n+\frac{1}{2\ell}} \\
	  & < D_{r,n} \frac{n+1}{\sqrt{c_r}}.
      \end{align*}
      Indeed, 
      \begin{itemize}[labelindent=0cm,leftmargin=*,topsep=0cm,partopsep=0cm,parsep=0cm,itemsep=0cm]
	  \item (a): The Gamma function has a global minima on the positive real line at $z_{min}\approx 1.46163$, it is strictly monotonically decreasing on $(0,z_{min})$ and strictly monotonically increasing on $(z_{min},\infty)$. The latter implies
	  \begin{align}
	      \Gamma(z_1) \le  \Gamma(z_2)\text{ for }z_{min}\le z_1 \le z_2.   \label{eq:Gamma:mon}
	  \end{align}
	  Let us choose $z_1 = \frac{rn+1}{2\ell} + \frac{r}{2\ell}$ and $z_2=\frac{rn+1}{2\ell} + \frac{2\ell}{2\ell}$.  $z_1 \le z_2$ since $r\le 2\ell$. With this choice
	  \eqref{eq:Gamma:mon} guarantees (a) with $D_{r,n}=1$ if
	    \begin{align*}
		z_{min} &\stackrel{(d)}{\le} \frac{n+2}{2\ell} = \left.\frac{rn+1}{2\ell} + \frac{r}{2\ell}\right|_{r=1} \le \\
			& \le \frac{rn+1}{2\ell} + \frac{r}{2\ell} = z_1. 
	    \end{align*}
	    If $n_s:=\ceil{2\ell z_{min} - 2} \le n$, then (d) holds. This means that (a) holds with
	    \begin{align*}
		D_{r,n} &= 1 \quad\text{ if } n_s \le n.
	    \end{align*}
	    For the remaining $n=2,\ldots,n_s-1$ values, (a) is fulfilled with equality using $D_{r,n} := \frac{\Gamma\left(\frac{rn+1}{2\ell} + \frac{2\ell}{2\ell}\right)}{\Gamma\left(\frac{rn+1}{2\ell} + \frac{r}{2\ell}\right)}$.
	  \item (b): We applied the $\Gamma(z+1) = z \Gamma(z)$ property.
	  \item (c): It follows from $r\le 2\ell$.
      \end{itemize}
\end{itemize}
To sum up, we got that 
    \begin{align*}
	  B_{r,n}  &\le \frac{D_{r,n}}{\sqrt{c_r}} (n+1),\text{ with}\\
	      D_{r,n} & = \begin{cases}
	                   1 & \text{if } n_s \le n\\
	                   \frac{\Gamma\left(\frac{rn+1}{2\ell} + \frac{2\ell}{2\ell}\right)}{\Gamma\left(\frac{rn+1}{2\ell} + \frac{r}{2\ell}\right)}  & n=2,\ldots,n_s-1
	                  \end{cases}.
      \end{align*}
      Thus, one can choose 
      \begin{align*}
	  K_r &=  \max\left(\frac{D_{r,2}}{\sqrt{c_r}},\ldots,\frac{D_{r,n_s-1}}{\sqrt{c_r}},\frac{1}{\sqrt{c_r}},1\right)
      \end{align*}
      in \eqref{eq:B-bound}.
\end{document}